\let\oldvec\vec
\documentclass[runningheads]{llncs}
\let\vec\oldvec

\usepackage{amsmath,amssymb} % define this before the line numbering.
\usepackage{graphicx}
\usepackage{color}
\usepackage{url}
\usepackage[width=122mm,left=12mm,paperwidth=146mm,height=193mm,top=12mm,paperheight=217mm]{geometry}

\begin{document}
% \linenumbers
\pagestyle{headings}
\mainmatter
\def\ECCV16SubNumber{9}  % Insert your submission number here

\title{A CNN Cascade for Landmark Guided Semantic Part Segmentation}

\titlerunning{A CNN Cascade for Landmark Guided Semantic Part Segmentation}
\authorrunning{A. Jackson, M. Valstar and G. Tzimiropoulos}

\author{Aaron S. Jackson, Michel Valstar, Georgios Tzimiropoulos}
\institute{School of Computer Science, The University of Nottingham,
  Nottingham, UK \\
  \texttt{\{aaron.jackson, michel.valstar,
    yorgos.tzimiropoulos\}@nottingham.ac.uk}}

\maketitle

\begin{abstract}
  This paper proposes a CNN cascade for semantic part segmentation
  guided by pose-specific information encoded in terms of a set of
  landmarks (or keypoints). There is large amount of prior work on
  each of these tasks separately, yet, to the best of our knowledge,
  this is the first time in literature that the interplay between pose
  estimation and semantic part segmentation is investigated. To
  address this limitation of prior work, in this paper, we propose a
  CNN cascade of tasks that firstly performs landmark localisation and
  then uses this information as input for guiding semantic part
  segmentation. We applied our architecture to the problem of facial
  part segmentation and report large performance improvement over the
  standard unguided network on the most challenging face
  datasets. Testing code and models will be published online at
  \url{http://cs.nott.ac.uk/~psxasj/}.  
\end{abstract}

\section{Introduction}

Pose estimation refers to the task of localising a set of landmarks
(or keypoints) on objects of interest like faces
\cite{cootes2001active}, the human body \cite{yang2011articulated} or
even birds \cite{zhang2015fine}. Locating these landmarks help
establish correspondences between two or more different instances of
the same object class which in turn has been proven useful for
fined-grained recognition tasks like face and activity
recognition. Part segmentation is a special case of semantic image
segmentation which is the task of assigning an object class label to
each pixel in the image. In part segmentation, the assigned label
corresponds to the part of the object that this pixel belongs to. In
this paper, we investigate whether pose estimation can guide
contemporary CNN architectures for semantic part segmentation. This
seems to be natural yet to the best of our knowledge this is the first
paper that addresses this problem. To this end, we propose a
Convolutional Neural Network (CNN) cascade for landmark guided part
segmentation and report large performance improvement over a standard
CNN for semantic segmentation that was trained without guidance.

Although the ideas and methods presented in this paper can probably be
applied to any structured deformable object (e.g. faces, human body,
cars, birds), we will confine ourselves to human faces. The main
reason for this is the lack of annotated datasets. To the best of our
knowledge, there are no datasets providing pixel-level annotation of
parts and landmarks at the same time. While this is also true for the
case of human faces, one can come up with pixel-level annotation of
facial parts by just appropriately connecting a pseudo-dense set of
facial landmarks for which many datasets and a very large number of
annotated facial images exist, see for example
\cite{sagonas2013semi}. Note that during testing we do not assume
knowledge of the landmarks' location, and what we actually show is
that a two-step process in which a CNN firstly predicts the landmarks
and then uses this information to segment the face largely outperforms
a CNN that was trained to directly perform facial part segmentation.

\begin{figure}
\centering
\includegraphics[width=\linewidth]{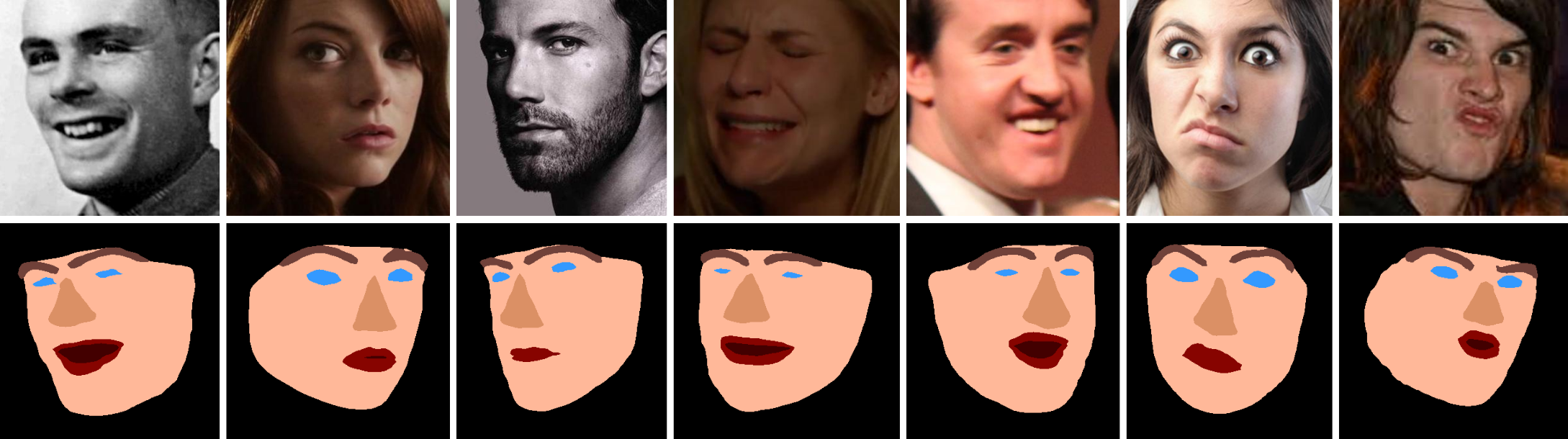}
\caption{Example faces and their corresponding output from the CNN
  cascade.}
\label{fig:sampler}
\end{figure}

\subsection{Main contributions}

In summary, this paper addresses the following research questions:
\begin{enumerate}
\item Is a CNN for facial part segmentation needed at all? One might
  argue that by just predicting the facial landmarks and then
  connecting them in the same way as we created the part labels, we
  could get high quality facial part segmentation thus completely
  by-passing the part segmentation task. Our first result in this
  paper is that indeed the latter method slightly outperforms a CNN
  trained for facial part segmentation (without guidance though).
\item Can facial landmarks be used for guiding facial part
  segmentation, thus reversing the result mentioned above? Indeed, we
  show that the proposed CNN cascade for landmark guided facial part
  segmentation largely outperforms both methods mentioned above
  without even requiring very accurate localisation of the
  landmarks. Some example output can be seen in Fig~\ref{fig:sampler}.
\end{enumerate}

\section{Related work}

This section reviews related work on semantic segmentation, facial
landmark localisation (also known as alignment) and facial part
segmentation.

\textbf{Face Alignment} State-of-the-art techniques in face alignment
are based on the so-called cascaded regression
\cite{dollar2010cascaded}. Given a facial image, such methods estimate
the landmarks' location by applying a sequence of regressors usually
learnt from SIFT \cite{lowe2004distinctive} or other hand-crafted
features. The regressors are learnt in a cascaded manner such that the
input to regressor $k$ is the estimate of the landmarks' location
provided by regressor $k-1$, see also
\cite{sanchez16,Cao2012shaperegression,xiongsupervised,zhu2015face,tzimiropoulos2015project}. The
first component in the proposed CNN cascade is a CNN landmark detector
based on VGG-16 \cite{simonyan2014very} converted to a fully
convolutional network \cite{long2015fully}. Although the main
contribution of our paper is not to propose a method for landmark
localisation, our CNN landmark localisation method performs comparably
with all aforementioned methods. One advantage of our method over
cascaded regression approaches is that it is not sensitive to
initialisation and hence it does not rely on accurate face detection.

\begin{figure}
\includegraphics[width=\linewidth]{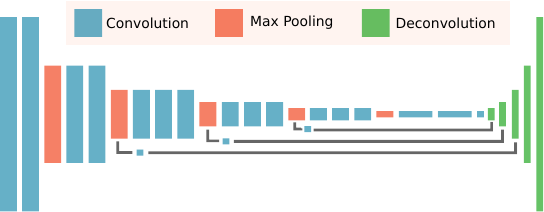}
\caption{Overview of the Fully Convolutional
  Network~\cite{long2015fully}, low level information providing
  refinement are reintroduced into the network during deconvolution.}
\label{fig:fcn}
\end{figure}

\textbf{Semantic Segmentation} Thanks to its ability to integrate
information from multiple CNN layers and its end-to-end training, the
Fully Convolutional Network (FCN) of \cite{long2015fully} has become
the standard basic component for all contemporary semantic
segmentation algorithms. The architecture of FCN is shown in
Fig. \ref{fig:fcn}. One of the limitations of the FCN is that
prediction is performed in low-resolution, hence a number of methods
have been recently proposed to compensate for this by usually applying
a Conditional Random Field (CRF) on top of the FCN output. The work of
\cite{chen2015semantic} firstly upsamples the predicted scores using
bilinear interpolation and then refines the output by applying a dense
CRF. The method of \cite{zheng2015conditional} performs recurrent
end-to-end training of the FCN and the dense CRF. Finally, the work in
\cite{noh2015learning} employs learnt deconvolution layers, as opposed
to fixing the parameters with an interpolation filter (as in
FCN). These filters learn to reconstruct the object's shape, instead
of just classifying each pixel. Although any of these methods could be
incorporated within the proposed CNN cascade, for simplicity, we used
the VGG-FCN \cite{simonyan2014very}. Note that all the aforementioned
methods perform unguided semantic segmentation, as opposed to the
proposed landmark-guided segmentation which incorporates information
about the pose of the object during both training and testing. To
encode pose specific information we augment the input to our
segmentation network with a multi-channel confidence map
representation using Gaussians centred at the predicted landmarks'
location, inspired by the human pose estimation method of
\cite{carreira2016human}. Note that \cite{carreira2016human} is
iterative an idea that could be also applied to our method, but
currently we have not observed performance improvement by doing so.

\textbf{Part Segmentation} There have been also a few works that
extend semantic segmentation to part segmentation with perhaps the
most well-known being the Shape Boltzman Machine
\cite{eslami2012generative,eslami2014shape}. This work has been
recently extended to incorporate CNN refined by CRF features (as in
\cite{chen2015semantic}) in \cite{tsogkas2015deep}. Note that this
work aims to refine the CNN output by applying a Restricted Boltzmann
Machine on top of it and does not make use of pose information as
provided by landmarks. In contrast, we propose an enhanced CNN
architecture which is landmark-guided, can be trained end-to-end and
yields large performance improvement without the need of further
refinement.

\textbf{Face Segmentation} One of the first face segmentation methods
prior to deep learning is known as LabelFaces
\cite{warrell2009labelfaces} which is based on patch classification
and further refinement via a hierarchical face model. Another
hierarchical approach to face segmentation based on Restricted
Boltzmann Machines was proposed in \cite{luo2012hierarchical}. More
recently, a multi-objective CNN has been shown to perform well for the
task of face segmentation in \cite{liu2015multi}. The method is based
on a CRF the unary and pairwise potentials of which are learnt via a
CNN. Softmax loss is used for the segmentation masks, and a logistic
loss is used to learn the edges. Additionally, the network makes use
of a non-parametric segmentation prior which is obtained as follows:
first facial landmarks on the test image are detected and then all
training images with most similar shapes are used to calculate an
average segmentation mask. This mask is finally used to augment
RGB. This segmentation mask might be blurry, does not encode pose
information and results in little performance improvement.

% accepts input of arbitrary size, and through layers of convolution
% the input is reduced to a deep of feature matrix. This is scored
% (per-pixel classification) and upsampled, rendering an unrefined but
% globally aware segmentation mask. To refine this, lower level
% (local) information is scored and incorporated. %% FCN

% An alternative approach to incorporating local information was
% demonstrated in~\cite{hariharan2015hypercolumns}. The output of each
% convolution is upsampled to the original input size. This allows a
% feature vector to be formed for each pixel, which can be used for
% labelling. To better capture the finer details of an image,
% \cite{noh2015learning} employs learnt deconvolution layers, as
% opposed to fixing the parameters with an interpolation filter, as in
% FCN. These filters learn to reconstruct the object's shape, instead
% of just classifying each pixel.

% Semantic Segmentation Guided by Detected Boundaries However... this
% is a bit boring because it seems they use it to mask the
% segmentation masks at the end? As opposed to what we do, which is
% pass this information to the network in the first conv layer.
% https://www.semanticscholar.org/paper/Object-Boundary-Guided-Semantic-Segmentation-Huang-Xia/79a3a2960945dcee46d0bb29944f4d7d1a97f37b/pdf

\section{Datasets}
\label{sec:dataset}

There are a few datasets which provide annotations of pixel-level
parts \cite{bo2011shape,kae2013augmenting,chen2014detect} but to the
best of our knowledge there are no datasets containing both part and
landmark annotations. Hence, in our paper we rely on datasets for
facial landmarking. These datasets provide a pseudo-dense set of
landmarks. Segmentation masks are constructed by joining the
groundtruth landmarks together to fully enclose each facial
component. The eyebrows are generated by a spline with a fixed width
relative to the normalised face size, to cover the entire eyebrow. The
selected classes are background, skin, eyebrows, eyes, nose, upper
lip, inner mouth and lower lip. While this results in straight edges
between landmarks, the network can learn a mean boundary for each
class. The output from the network will be actually smoother than the
groundtruth.

\begin{figure}
\centering
\includegraphics[height=3.5cm]{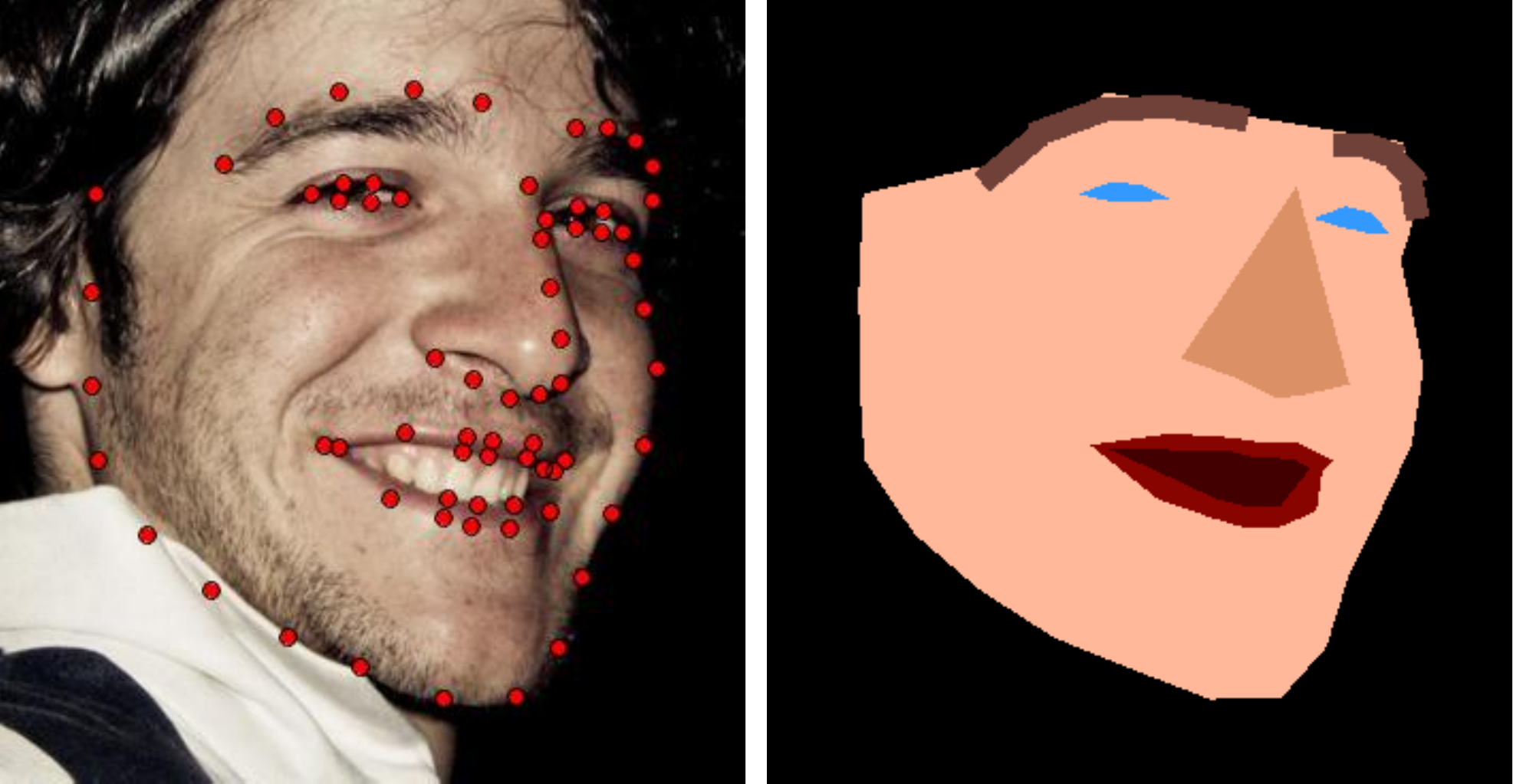}
\hspace{-1.4mm}
\includegraphics[height=3.5cm]{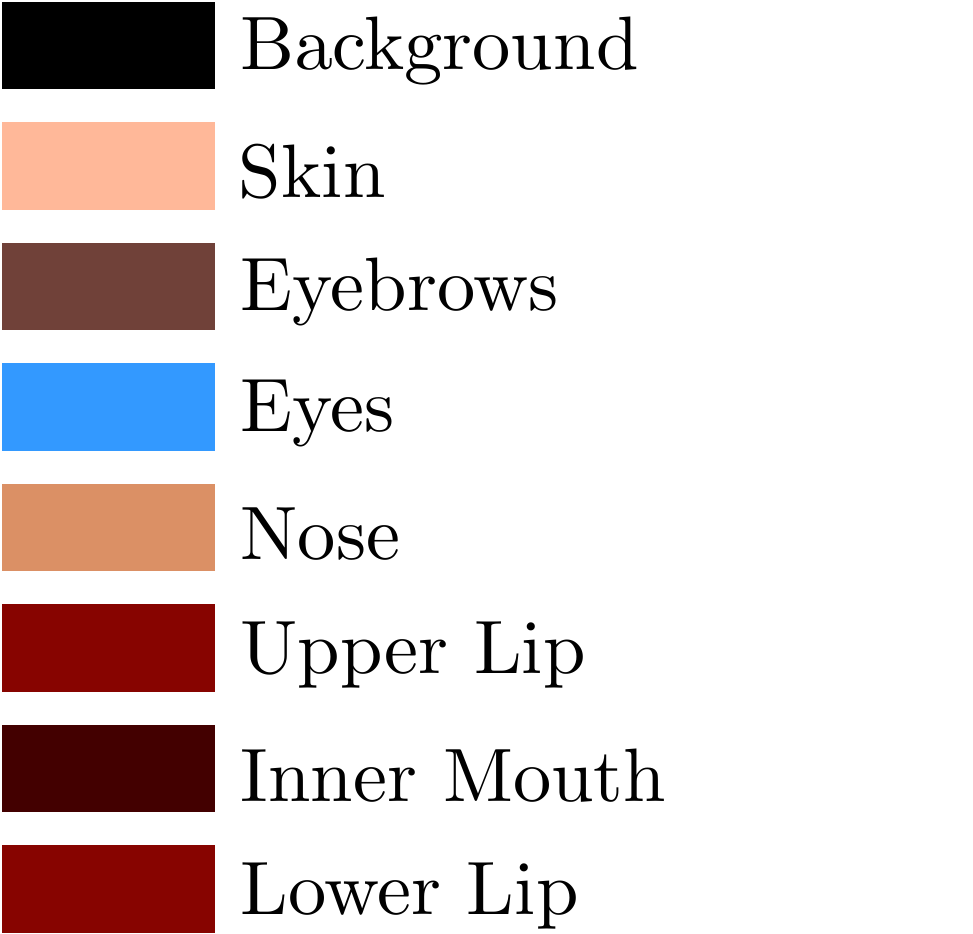}
\caption{Example groundtruth segmentation mask produced from the
  groundtruth landmarks.}
\label{fig:gtmasks}
\end{figure}

This process is illustrated in Fig.~\ref{fig:gtmasks}.

For our experiments we used the 68-point landmark annotations provided
by the 300W challenge \cite{sagonas2013300}. In particular the
training sets of LFPW \cite{belhumeur2011localizing}, Helen
\cite{le2012interactive}, AFW \cite{ramanan2011} and iBUG
\cite{sagonas2013300} are all used for training while the 300W test
set (600 images) is used for testing. Both training and test sets
contain very challenging images in terms of appearance, pose,
expression and occlusion.

This collection of images undergoes some pre-processing before they
are used to train the network. The faces are normalised to be of equal
size and cropped with some noise added to the position of the bounding
box. Not all images are the same size, but their height is fixed at
350 pixels. With probability $p=0.5$, a randomly sized black
rectangle, large enough to occlude an entire component is layered over
the input image. This assists the network in learning a robustness to
partial occlusion.

\section{Method}
\label{sec:proposed}

\begin{figure}
\includegraphics[width=\linewidth]{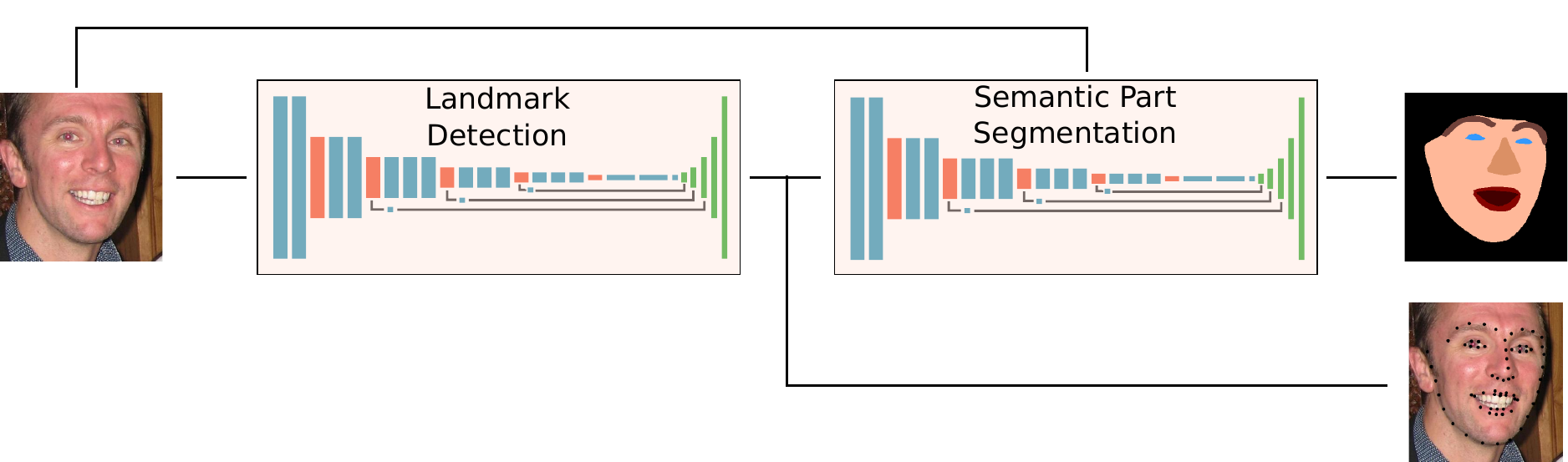}
\caption{The proposed architecture, comprising of two separate Fully
  Convolutional Networks. The first performs Landmark Detection, the
  output of which is encoded as multichannel representation which is
  then passed into the Semantic Part Segmentation network.}
\label{fig:proposed}
\end{figure}

We propose a CNN cascade (shown in Fig.~\ref{fig:proposed} and listed
in Table~\ref{tab:archlist}) which performs landmark localisation
followed by facial part segmentation. Our cascade was based on the
VGG-FCN \cite{simonyan2014very,long2015fully} using Caffe
\cite{jia2014caffe} and consists of two main components:
\begin{enumerate}
\item Firstly, an FCN is trained to detect facial landmarks using
  Sigmoid Cross Entropy Loss.
\item Secondly, inspired by the human pose estimation method of
  \cite{carreira2016human}, the detected 68 landmarks are encoded as
  68 separate channels each of which contains a 2D Gaussian centred
  at the corresponding landmark's location. The 68 channels are then
  stacked along with the original image and passed into our
  segmentation network. This is a second FCN trained for facial part
  segmentation using as input the stacked representation of 2D
  Gaussians and image, and a standard Softmax loss.
\end{enumerate}

Overall we encode pose specific information by augmenting the input to
our segmentation network with a multi-channel confidence map
representation using Gaussians centred at the predicted landmarks'
location. Hence, our FCN for semantic segmentation is trained to
produce high quality, refined semantic masks by incorporating low
level information with globally aware information. Each of the
aforementioned components is now discussed in more detail:

% Training FCN for unguided part segmentation
% % learning rate, momentum, number of iterations, training steps
% % chosen loss
% Training FCN for landmark localisation
% % learning rate, momentum, number of iterations, training steps,
% % chosen loss
\textbf{Facial Landmark Detection} The training procedure for landmark
detection is similar to training FCN for part segmentation. Landmarks
are encoded as 2D Gaussians centred at the provided landmarks'
location. Each landmark is allocated its own channel to prevent
overlapping with other landmarks and allow the network to more easily
distinguish between each point. The main difference with part
segmentation is the loss function. Sigmoid Cross Entropy Loss
\cite{zhang2015fine} was chosen to regress the likelihood of a pixel
containing a point. More concretely, given our groundtruth Gaussians
$\hat{p}$ and predicted Gaussians $p$, each of equal dimensions
$ N \times W \times H$, we can define the Sigmoid Cross Entropy loss
$l$ as follows:

\[
  l = \frac{1}{N} \sum^{N}_{n=1} \sum^{W}_{i=1} \sum^{H}_{j=1}
  [p^n_{i,j} \log(\hat{p}^n_{i,j}) + (1 -p^n_{i,j})
  \log (1-\hat{p}^n_{i,j}) ].
\]

The loss was scaled by $1\mathrm{e}^{-5}$ and a learning rate of
0.0001 was used. The network was trained in steps as previously
described, for approximately 400,000 iterations, until convergence.

% Finetuning part segmentation to become guided
\textbf{Guided Facial Part Segmentation} To train our guided FCN part
segmentation network we followed \cite{long2015fully}. Softmax Loss
was also used. If $N$ is the number of outputs (in our case, classes),
$p_{i,j}$ is the predicted output for pixel $(i,j)$, and $n$ is the
true label for pixel $(i,j)$, then the Softmax loss $l$ can be defined
as:

% \[
% l = \frac{-1}{N} \sum^{N}_{n=1} \log ( p^{n}_{i,j} )
% \]

\[
l = \frac{-1}{N} \sum^{W}_{i=1} \sum^{H}_{j=1} \log(p^n_{i,j}).
\]

We firstly trained an unguided FCN for facial part segmentation
following \cite{long2015fully}. Initially, the network was trained as
32 stride, where no information from the lower layers is used to
refine the output. This followed by introducing information from
pool4, followed by pool3. A learning rate of 0.0001 was chosen, and a
momentum of 0.9. The network was trained for approximately 300,000
iterations until convergence.

Then, our guided FCN was initialised from the weights of the unguided
one, by expanding the first layer to accommodate the additional 68
input channels. As mentioned earlier, each channel contains a 2D
Gaussian centred at the corresponding landmark's location.  A key
aspect of our cascade is how the landmarks' location is determined
during training. We cannot use the groundtruth landmark locations nor
the prediction of our facial landmark detection network on our
training set as those will be significantly more accurate than those
observed during testing. Hence, we applied our facial landmark
detection network on our validation set and recorded the landmark
localisation error. We used this error to create a multivariate
Gaussian noise model that was added to the groundtruth landmark
locations of our training set. This way our guided segmentation
network was initialised with much more realistic input in terms of
landmarks' location. Furthermore, the same learning rate of 0.0001 was
used. For the first 10,000 iterations, training was disabled on all
layers except for the first. This allowed the network to warm up
slightly, and prevent the parameters in other layers from getting
destroyed by a high loss.

\begin{table}
  \caption{The
    VGG-FCN~\cite{simonyan2014very,long2015fully}
    architecture employed by our landmark detection and semantic
    part segmentation network.}
\label{tab:archlist}
\centering
\begin{tabular}{|c|c|c|c|}
\hline
Layer Name           & Kernel       & Stride       &  Outputs  \\
\hline\hline
\texttt{conv1\_1}    & $3 \times 3$ & $1 \times 1$ &  64  \\
\texttt{conv1\_2}    & $3 \times 3$ & $1 \times 1$ &  64  \\
\texttt{pool1}       & $2 \times 2$ & $2 \times 2$ &  --  \\
\texttt{conv2\_1}    & $3 \times 3$ & $1 \times 1$ &  128 \\
\texttt{conv2\_2}    & $3 \times 3$ & $1 \times 1$ &  128 \\
\texttt{pool2}       & $2 \times 2$ & $2 \times 2$ &  --  \\
\texttt{conv3\_1}    & $3 \times 3$ & $1 \times 1$ &  256 \\
\texttt{conv3\_2}    & $3 \times 3$ & $1 \times 1$ &  256 \\
\texttt{conv3\_3}    & $3 \times 3$ & $1 \times 1$ &  256 \\
\texttt{pool3}       & $2 \times 2$ & $2 \times 2$ &  --  \\
\texttt{conv4\_1}    & $3 \times 3$ & $1 \times 1$ & 512  \\
\texttt{conv4\_2}    & $3 \times 3$ & $1 \times 1$ & 512  \\
\texttt{conv4\_3}    & $3 \times 3$ & $1 \times 1$ & 512  \\

\hline
\end{tabular}
\begin{tabular}{|c|c|c|c|}
\hline
Layer Name           & Kernel       & Stride       &  Outputs  \\
\hline\hline
\texttt{pool4}       & $2 \times 2$ & $2 \times 2$ & --   \\
\texttt{conv5\_1}    & $3 \times 3$ & $1 \times 1$ & 512  \\
\texttt{conv5\_2}    & $3 \times 3$ & $1 \times 1$ & 512  \\
\texttt{conv5\_3}    & $3 \times 3$ & $1 \times 1$ & 512  \\
\texttt{pool5}       & $2 \times 2$ & $2 \times 2$ & --   \\
\texttt{fc6\_conv}   & $7 \times 7$ & $1 \times 1$ & 4096 \\
\texttt{fc7\_conv}   & $1 \times 1$ & $1 \times 1$ & 4096 \\
\texttt{fc8\_conv}   & $1 \times 1$ & $1 \times 1$ & 68 or 7 \\
\texttt{deconv\_32}   & $4 \times 4$ & $2 \times 2$ & 68 or 7 \\
\texttt{score\_pool4} & $1 \times 1$ & $1 \times 1$ & 68 or 7 \\
\texttt{deconv\_16}   & $4 \times 4$ & $2 \times 2$ & 68 or 7 \\
\texttt{score\_pool3} & $1 \times 1$ & $1 \times 1$ & 68 or 7 \\
\texttt{deconv\_8}    & $16 \times 16$ & $8 \times 8$ & 68 or 7 \\
\hline
\end{tabular}
\end{table}

\section{Experiments}

\subsection{Overview of Results}

In all experiments we used the training and test sets detailed in
Section \ref{sec:dataset}. As a performance measure, we used the
familiar intersection over union measure \cite{long2015fully}. We
report a comparison between the performance of four different methods
of interest:
\begin{enumerate}
\item The first method is the VGG-FCN trained for facial part
  segmentation. We call this method \textbf{Unguided}.
\item The second method is the part segmentation result obtained by
  joining the landmarks obtained from VGG-FCN trained for facial
  landmark detection. We call this method \textbf{Connected
    Landmarks}.
\item The third method is the proposed landmark guided part
  segmentation network where the input is the groundtruth landmarks'
  location. We call this method \textbf{Guided by Groundtruth}.
\item Finally, the fourth method is the proposed landmark guided part
  segmentation network when input is detected landmarks' location. We
  call this method \textbf{Guided by Detected}.
\end{enumerate}
The first two methods are the baselines in our experiments while the
third one provides an upper bound in performance. The fourth method is
the proposed CNN cascade.

% \subsection{Facial Part Segmentation from Detected Landmarks}

% In this experiment we attempt to show the performance of facial part
% segmentation by connecting the detected landmarks. Although
% groundtruth landmarks were used to generate the groundtruth
% segmentation masks for training, the quality of part segmentation by
% joining detected landmarks was quite disappointing.

\subsection{Unguided Facial Part Segmentation}

\begin{figure}
\includegraphics[width=0.5\linewidth]{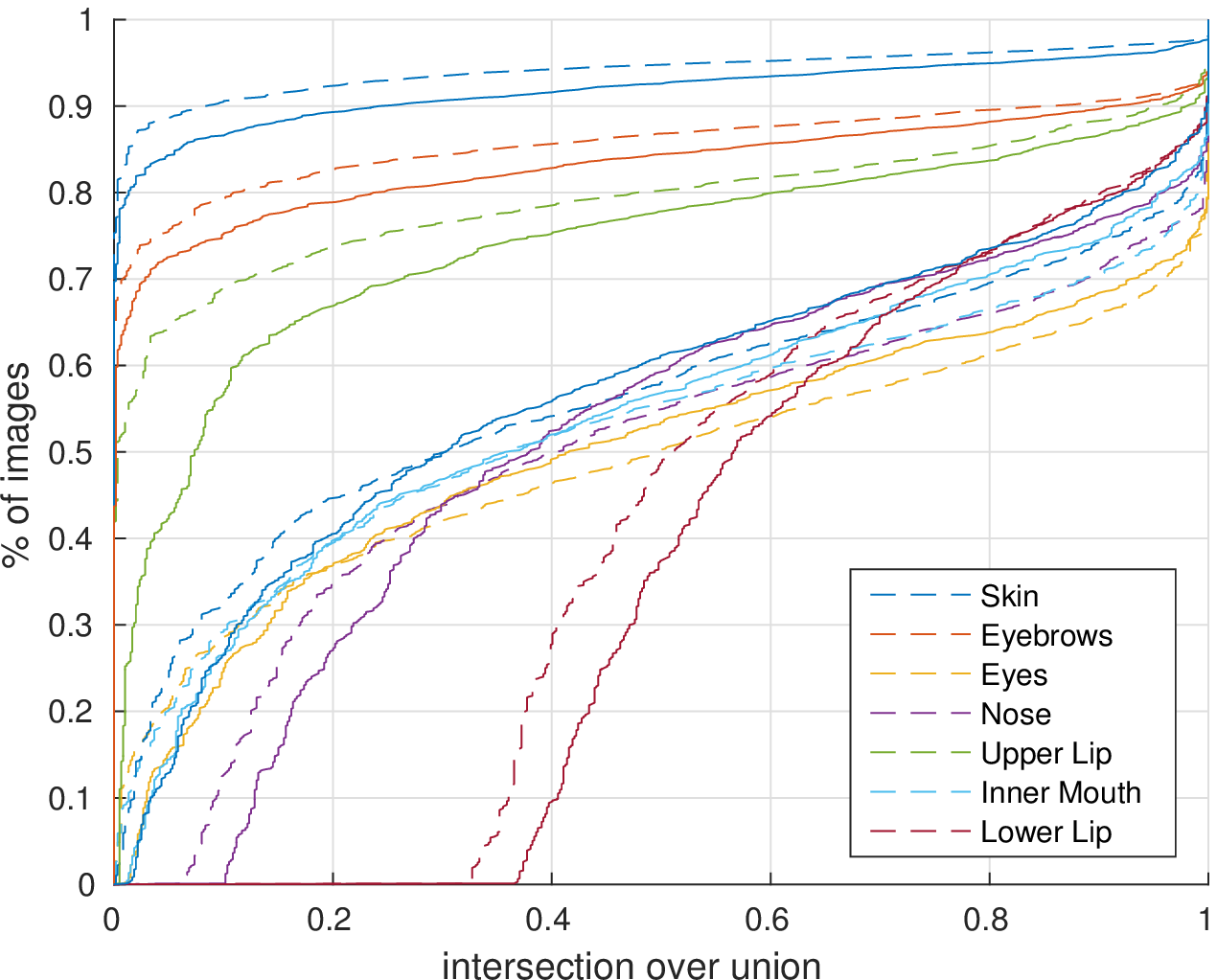}
\includegraphics[width=0.5\linewidth]{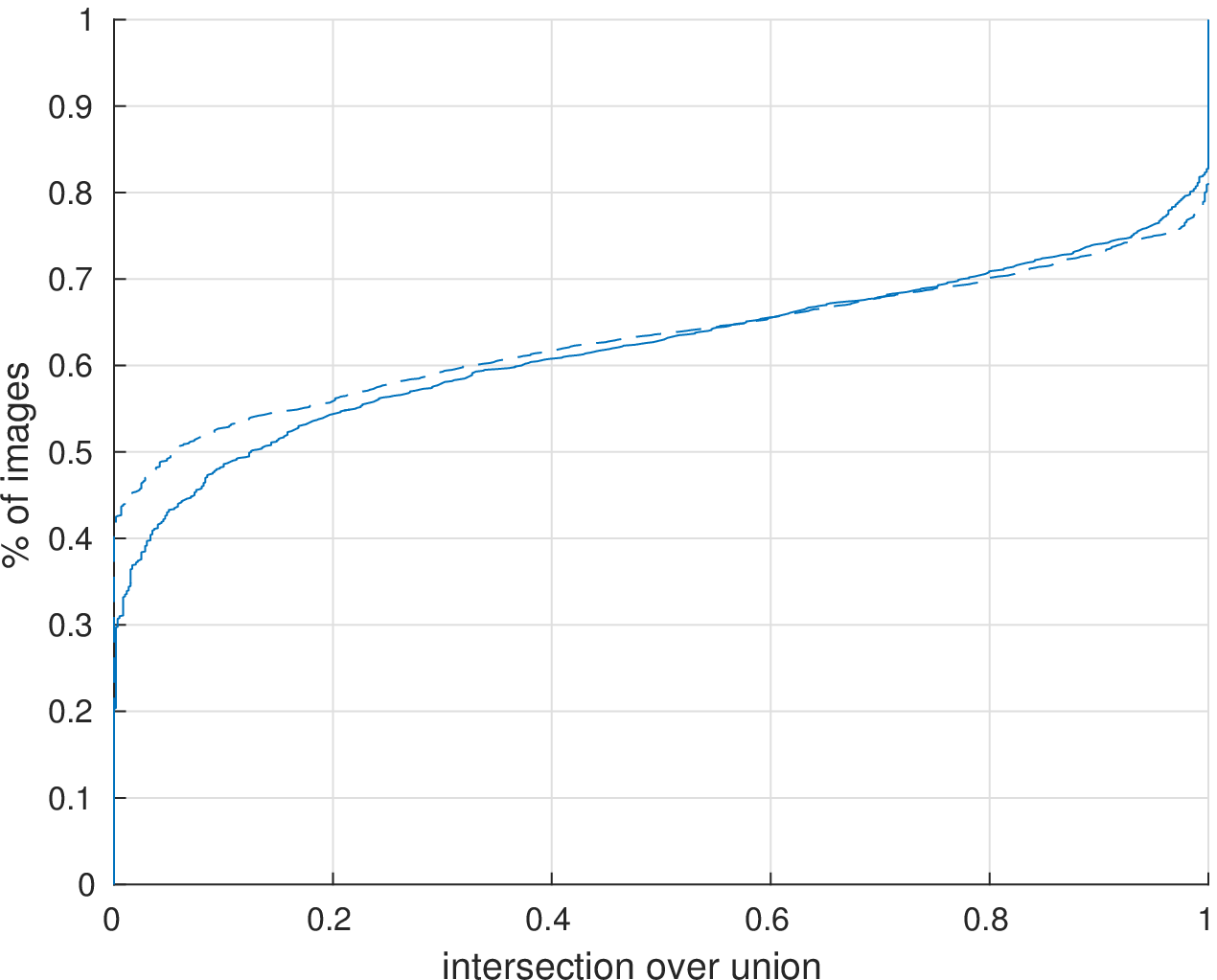}
\caption{Comparison of Unguided (---) and Connected Landmarks
(-{ }-). Per-class averages shown on the right.}
\label{fig:lan-vs-un}
\end{figure}

To establish a baseline, an unguided fully convolutional network was
firstly trained. This was done as described in the FCN paper
\cite{long2015fully} and Section \ref{sec:proposed}. Some visual
results can be seen in Fig.~\ref{fig:visual}. Additionally, a second
baseline was obtained by simply connecting the landmarks of our facial
landmark detection network also described in Section
\ref{sec:proposed}. The performance of both baselines can be seen in
Fig. \ref{fig:lan-vs-un}. We may observe that connecting the landmarks
appears to offer slightly better performance than FCN for part
segmentation alone. Nevertheless, we need to emphasise that the
groundtruth masks were obtained by connecting the landmarks and hence
there is some bias towards the connecting the landmarks approach.

\subsection{Guided Facial Part Segmentation with Groundtruth}

\begin{figure}
\includegraphics[width=0.5\linewidth]{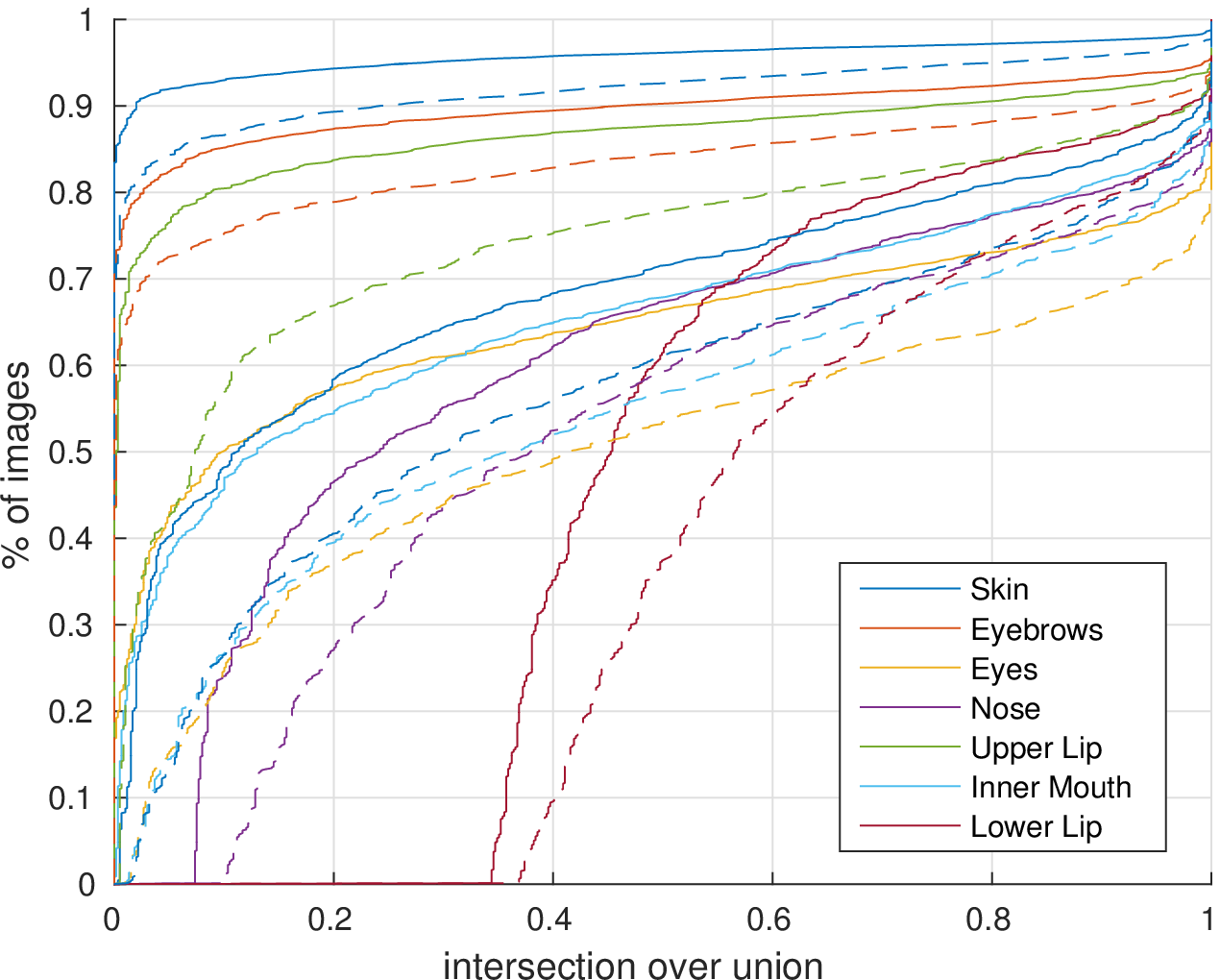}
\includegraphics[width=0.5\linewidth]{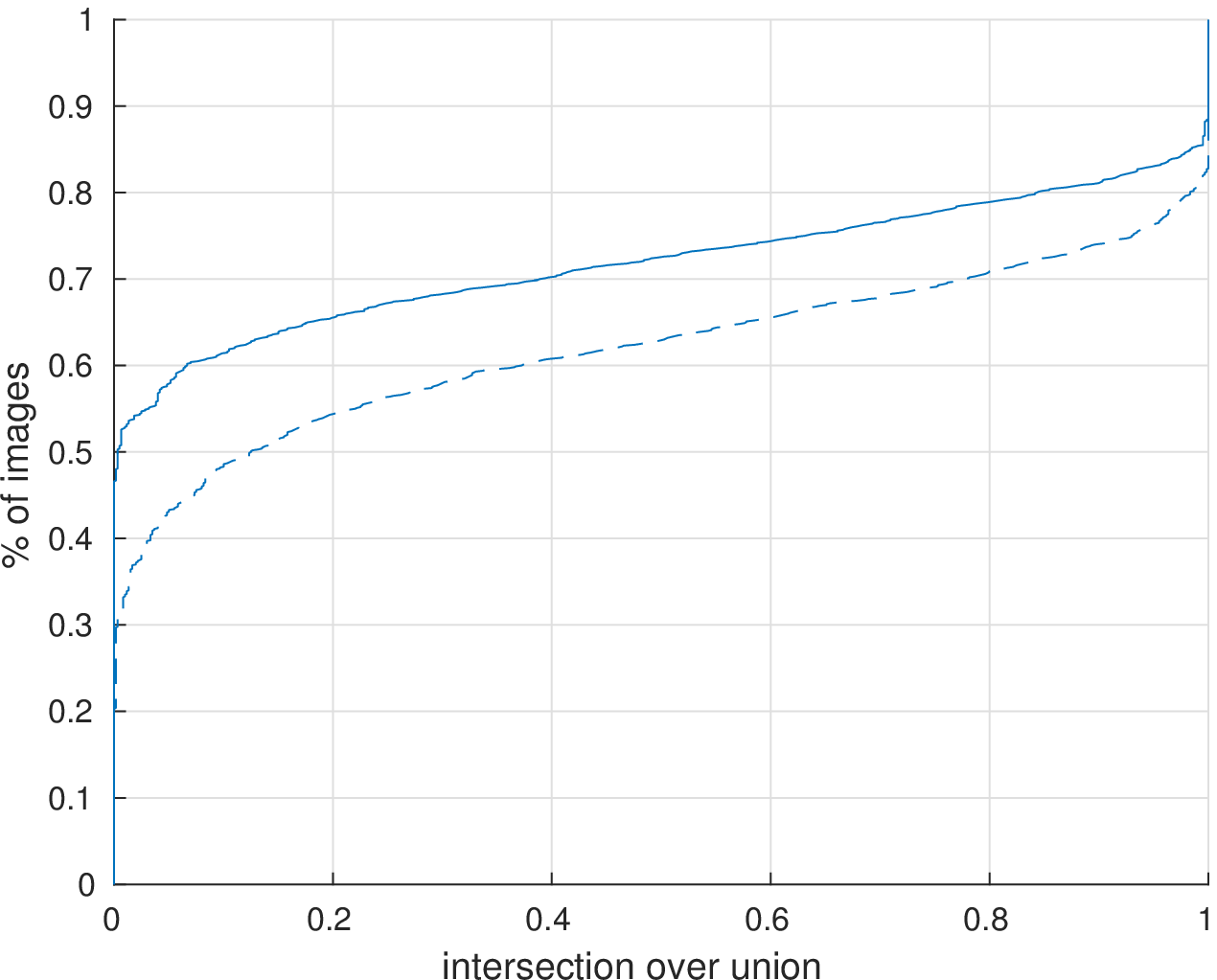}
\caption{Comparison of guided by groundtruth (---) and unguided (-{ }-)
facial part segmentation. Per-class averages shown on the right.
}
\label{fig:gt-vs-un}
\end{figure}

To establish an upper bounds to our performance, a fully convolutional
network was trained to accept guidance from groundtruth landmarks. As
described in Section~\ref{sec:proposed}, the guidance is provided in
the form of landmarks encoded as 2D Gaussians. The performance
difference between unguided and groundtruth guided part segmentation
can be seen in Fig.~\ref{fig:gt-vs-un}. As we may observe the
difference in performance between the two methods is huge. These
results are not surprising given that the groundtruth semantic masks
are generated from the landmarks guiding the network. Furthermore,
landmark detection offers an advantage because, in the case of faces,
there can only be one tip of the nose, and one left side of the
mouth. Giving some information to the network about where it is likely
to be located can offer a significant advantage. Our next experiment
shows that this is still the case when detected landmarks are used
instead of groundtruth landmarks.

\subsection{Guided Facial Part Segmentation with Detected Landmarks}

% Maybe this should be guidance by detected vs. unguided?
\begin{figure}
\includegraphics[width=0.5\linewidth]{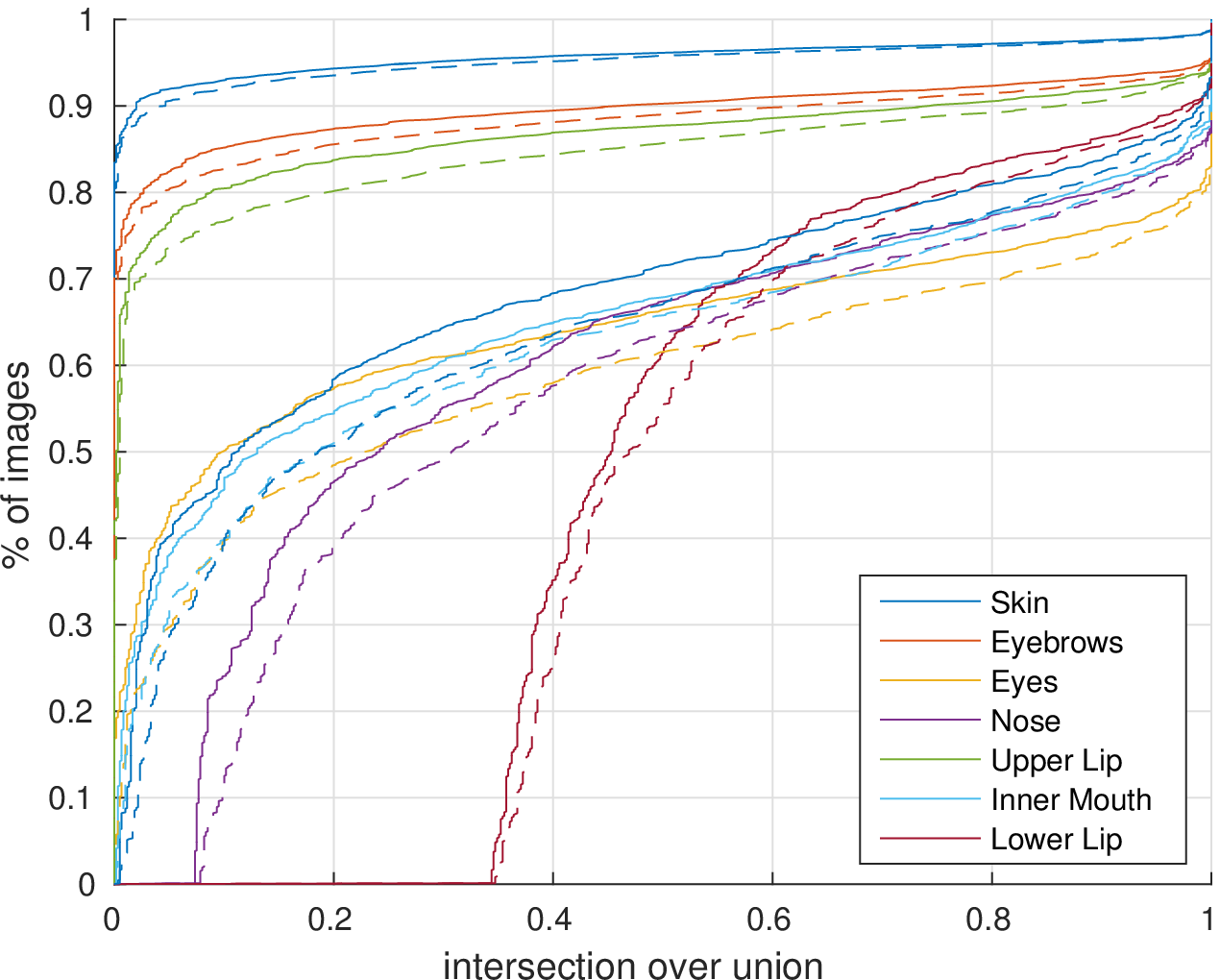}
\includegraphics[width=0.5\linewidth]{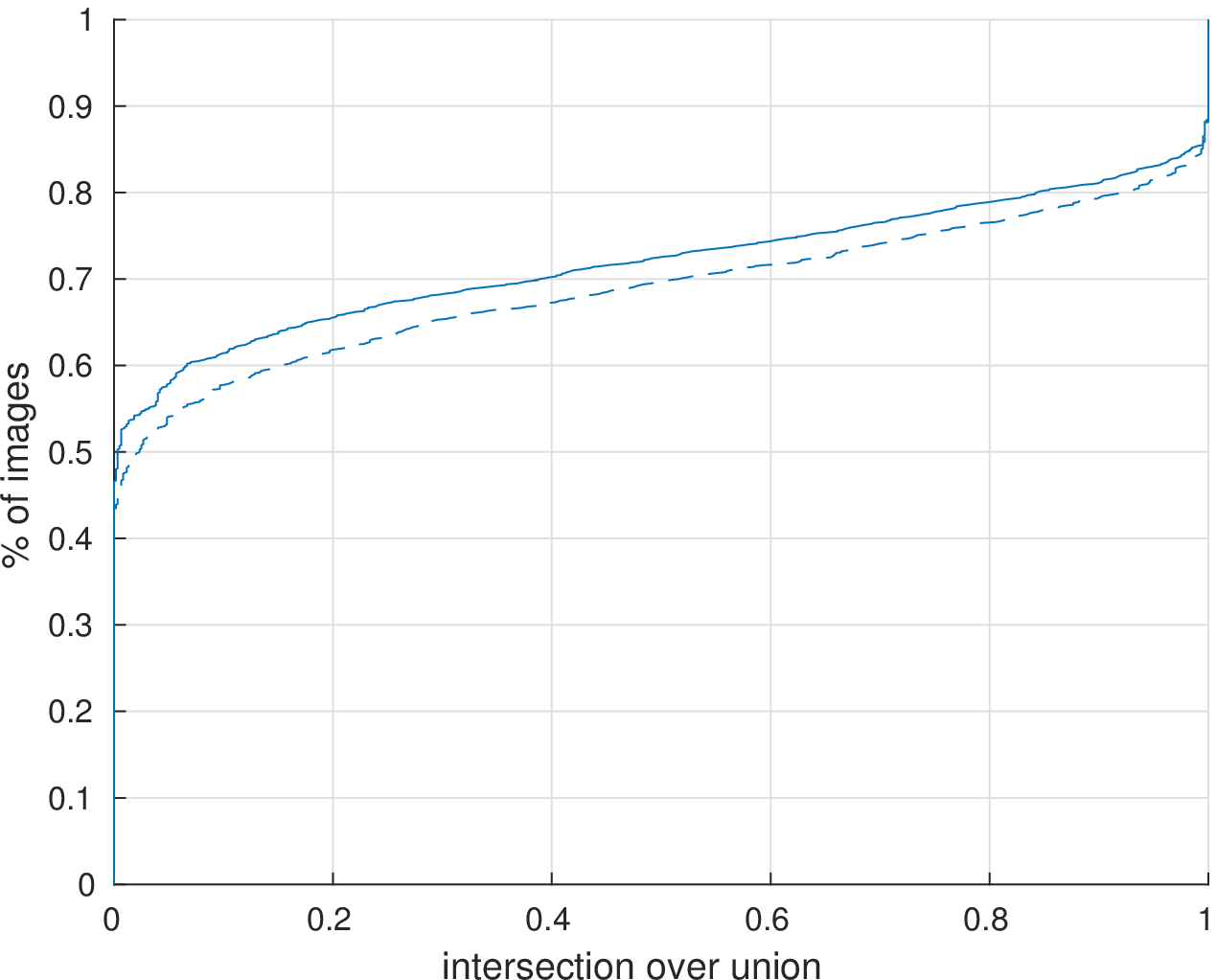}
\caption{Comparison of guidance from groundtruth landmarks (---) and
  guidance from detected landmarks (-{ }-). Per-class averages shown
  on the right.  }
\label{fig:gt-vs-det}
\end{figure}

\begin{figure}
\includegraphics[width=\linewidth]{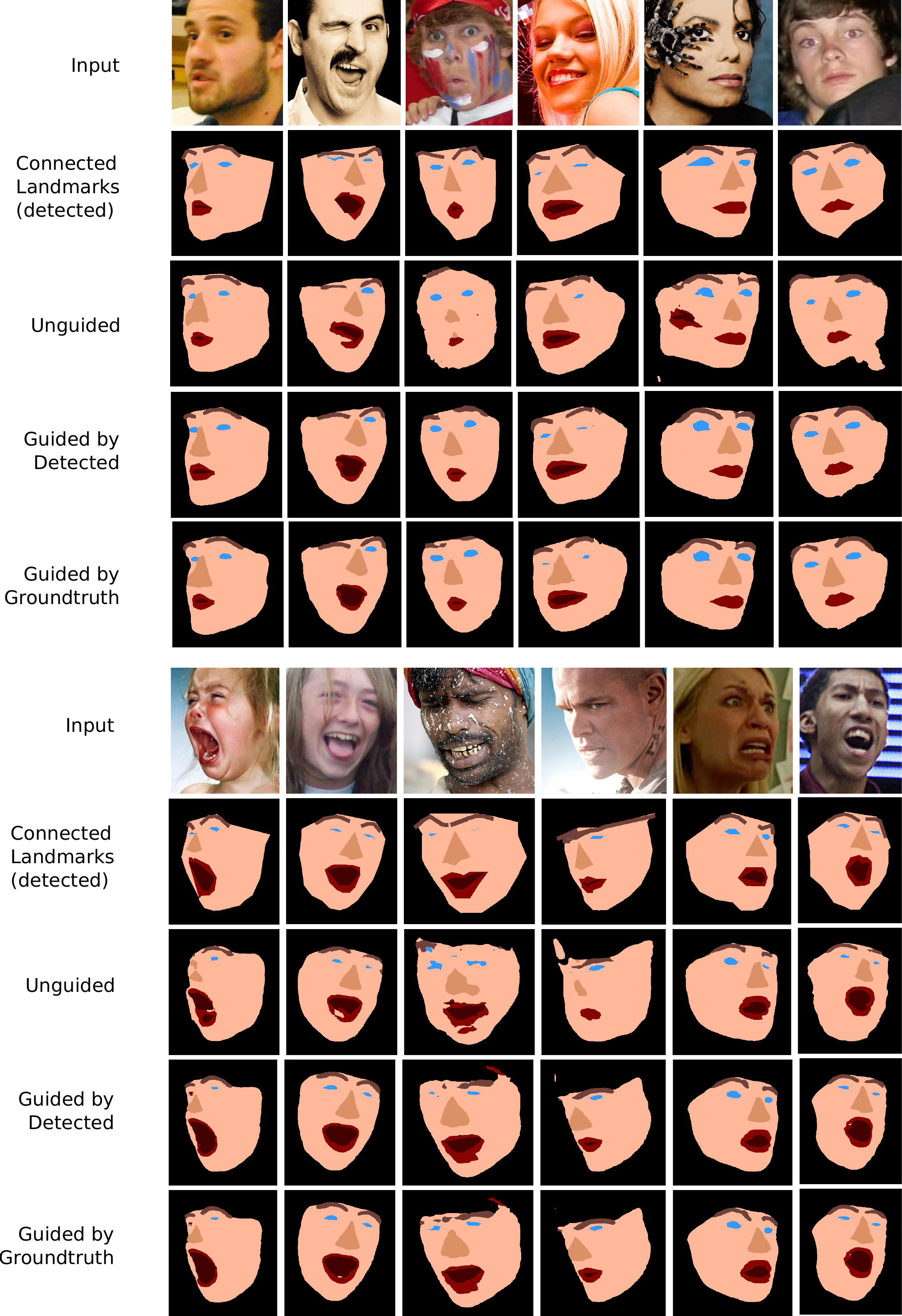}
\caption{Some visual results showing where the unguided network begins
  to fail, and where the guidance begins to pay off. Observe how
  visually close the results of the guided by groundtruth landmarks
  and the guided by detected landmarks networks are.}
\label{fig:visual}
\end{figure}

\begin{figure}
\centering
\includegraphics[width=0.6\linewidth]{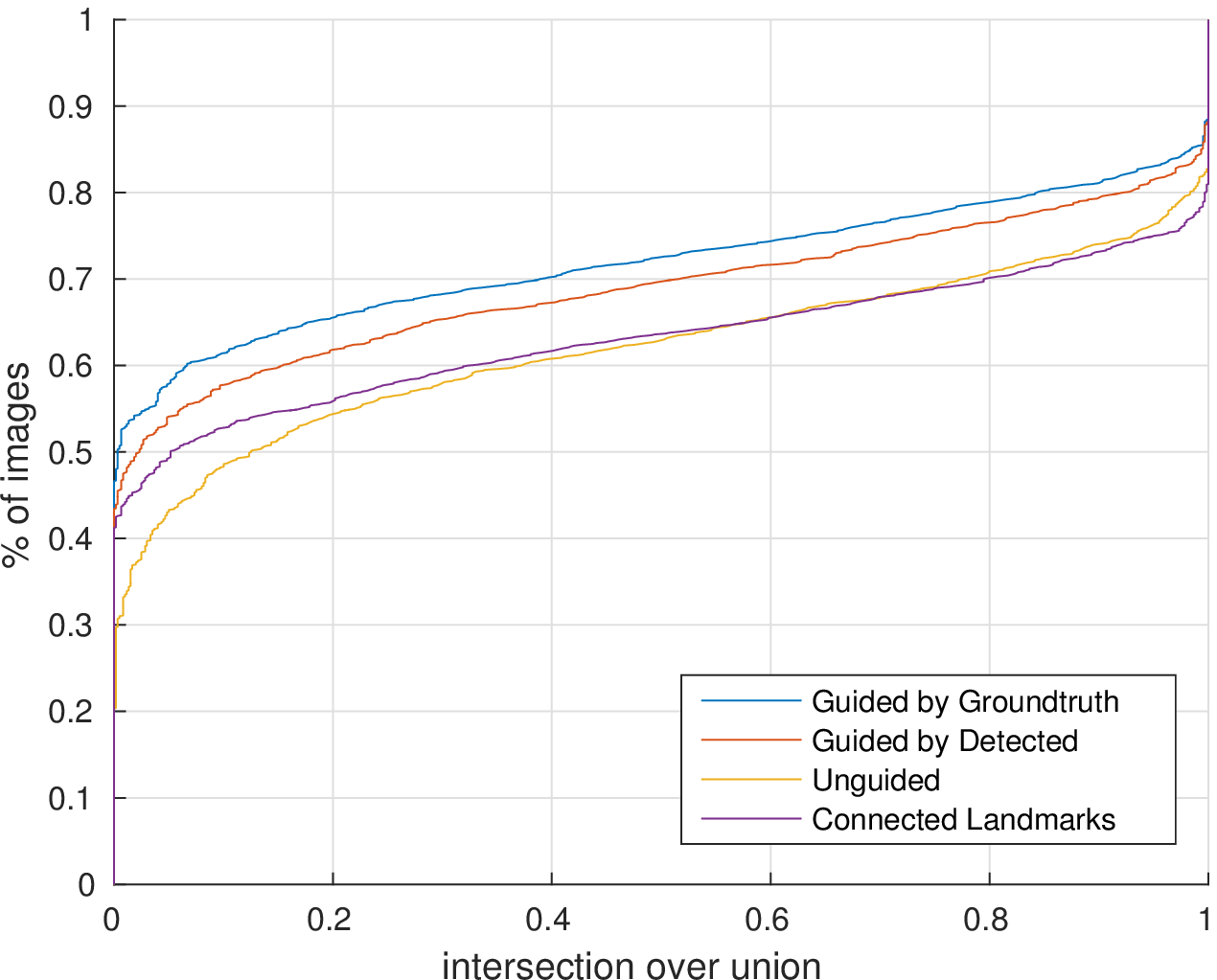}
\caption{Average performance of the four tested methods over all
  facial components: part segmentation guided by groundtruth
  landmarks, part segmentation guided by detected landmarks, unguided
  part segmentation, part segmentation by joining up the detected
  landmarks.}
\label{fig:all}
\end{figure}

With our upper bound and baselines defined, we can now see how much of
an improvement we can achieve by guiding the network with our
detected landmarks. The output of the landmark detection network is
passed into the part segmentation network along with the original
input image. We acknowledge that the performance of our landmark
detector is far from groundtruth. We measure the performance as mean
point to point Euclidean distance normalised by the outer interocular
Euclidean distance, as in~\cite{sagonas2013300}. This results in an
error of $0.0479$. However, we show that the performance of the
segmentation is improved significantly. The results of facial part
segmentation guided by the detected landmarks, compared to the network
guided by groundtruth landmarks can be seen in
Fig~\ref{fig:gt-vs-det}. Our main result is that performance of the
guided by detected network is very close to the that of the guided by
groundtruth illustrating that in practice accurate landmark
localisation is not really required to guide segmentation. Some visual
results can be seen in Fig.~\ref{fig:visual}. Also, performance over
all components for all methods is given in Fig.~\ref{fig:all}.

\section{Conclusion}

In this paper we proposed a CNN architecture to improve the
performance of part segmentation by task delegation. In doing so, we
provided both landmark localisation and semantic part segmentation on
human faces. However, our method should be applicable to our objects
as well. This is the focus of our ongoing work. We are also looking
into how the segmentation masks can be further used to improve
landmark localisation accuracy, thus leading to a recurrent
architecture. Future work may also compare the performance of this
method with a multitask architecture.

\section*{Acknowledgements}

Aaron Jackson was funded by a PhD scholarship from the University of
Nottingham. The work of Valstar is also funded by European Union
Horizon 2020 research and innovation programme under grant agreement
number 645378. Georgios Tzimiropoulos was supported in part by the
EPSRC project EP/M02153X/1 Facial Deformable Models of Animals.

\clearpage
\bibliographystyle{splncs}
\bibliography{egbib}

\begin{thebibliography}{10}

\bibitem{cootes2001active}
Cootes, T., Edwards, G., Taylor, C.:
\newblock Active appearance models.
\newblock TPAMI \textbf{23}(6) (2001)  681--685

\bibitem{yang2011articulated}
Yang, Y., Ramanan, D.:
\newblock Articulated pose estimation with flexible mixtures-of-parts.
\newblock In: CVPR. (2011)

\bibitem{zhang2015fine}
Zhang, N., Shelhamer, E., Gao, Y., Darrell, T.:
\newblock Fine-grained pose prediction, normalization, and recognition.
\newblock arXiv preprint arXiv:1511.07063 (2015)

\bibitem{sagonas2013semi}
Sagonas, C., Tzimiropoulos, G., Zafeiriou, S., Pantic, M.:
\newblock A semi-automatic methodology for facial landmark annotation.
\newblock In: CVPR-W. (2013)

\bibitem{dollar2010cascaded}
Doll{\'a}r, P., Welinder, P., Perona, P.:
\newblock Cascaded pose regression.
\newblock In: CVPR. (2010)

\bibitem{lowe2004distinctive}
Lowe, D.G.:
\newblock Distinctive image features from scale-invariant keypoints.
\newblock IJCV \textbf{60}(2) (2004)  91--110

\bibitem{sanchez16}
E.~Sánchez-Lozano, B.~Martinez, G.T., Valstar, M.:
\newblock Cascaded continuous regression for real-time incremental face
  tracking.
\newblock In: European Conference on Computer Vision. (2016)

\bibitem{Cao2012shaperegression}
Cao, X., Wei, Y., Wen, F., Sun, J.:
\newblock Face alignment by explicit shape regression.
\newblock In: CVPR. (2012)

\bibitem{xiongsupervised}
Xiong, X., De~la Torre, F.:
\newblock Supervised descent method and its applications to face alignment.
\newblock In: CVPR. (2013)

\bibitem{zhu2015face}
Zhu, S., Li, C., Change~Loy, C., Tang, X.:
\newblock Face alignment by coarse-to-fine shape searching.
\newblock In: CVPR. (2015)

\bibitem{tzimiropoulos2015project}
Tzimiropoulos, G.:
\newblock Project-out cascaded regression with an application to face
  alignment.
\newblock In: CVPR. (2015)

\bibitem{simonyan2014very}
Simonyan, K., Zisserman, A.:
\newblock Very deep convolutional networks for large-scale image recognition.
\newblock arXiv preprint arXiv:1409.1556 (2014)

\bibitem{long2015fully}
Long, J., Shelhamer, E., Darrell, T.:
\newblock Fully convolutional networks for semantic segmentation.
\newblock In: CVPR. (2015)

\bibitem{chen2015semantic}
Chen, L.C., Papandreou, G., Kokkinos, I., Murphy, K., Yuille, A.L.:
\newblock Semantic image segmentation with deep convolutional nets and fully
  connected crfs.
\newblock In: ICLR. (2015)

\bibitem{zheng2015conditional}
Zheng, S., Jayasumana, S., Romera-Paredes, B., Vineet, V., Su, Z., Du, D.,
  Huang, C., Torr, P.H.:
\newblock Conditional random fields as recurrent neural networks.
\newblock In: CVPR. (2015)

\bibitem{noh2015learning}
Noh, H., Hong, S., Han, B.:
\newblock Learning deconvolution network for semantic segmentation.
\newblock In: Proceedings of the IEEE International Conference on Computer
  Vision. (2015)  1520--1528

\bibitem{carreira2016human}
Carreira, J., Agrawal, P., Fragkiadaki, K., Malik, J.:
\newblock Human pose estimation with iterative error feedback.
\newblock In: CVPR. (2016)

\bibitem{eslami2012generative}
Eslami, S., Williams, C.:
\newblock A generative model for parts-based object segmentation.
\newblock In: NIPS. (2012)

\bibitem{eslami2014shape}
Eslami, S.A., Heess, N., Williams, C.K., Winn, J.:
\newblock The shape boltzmann machine: a strong model of object shape.
\newblock IJCV \textbf{107}(2) (2014)  155--176

\bibitem{tsogkas2015deep}
Tsogkas, S., Kokkinos, I., Papandreou, G., Vedaldi, A.:
\newblock Deep learning for semantic part segmentation with high-level
  guidance.
\newblock arXiv preprint arXiv:1505.02438 (2015)

\bibitem{warrell2009labelfaces}
Warrell, J., Prince, S.J.:
\newblock Labelfaces: Parsing facial features by multiclass labeling with an
  epitome prior.
\newblock In: Image Processing (ICIP), 2009 16th IEEE International Conference
  on, IEEE (2009)  2481--2484

\bibitem{luo2012hierarchical}
Luo, P., Wang, X., Tang, X.:
\newblock Hierarchical face parsing via deep learning.
\newblock In: Computer Vision and Pattern Recognition (CVPR), 2012 IEEE
  Conference on, IEEE (2012)  2480--2487

\bibitem{liu2015multi}
Liu, S., Yang, J., Huang, C., Yang, M.H.:
\newblock Multi-objective convolutional learning for face labeling.
\newblock In: Proceedings of the IEEE Conference on Computer Vision and Pattern
  Recognition. (2015)  3451--3459

\bibitem{bo2011shape}
Bo, Y., Fowlkes, C.C.:
\newblock Shape-based pedestrian parsing.
\newblock In: Computer Vision and Pattern Recognition (CVPR), 2011 IEEE
  Conference on, IEEE (2011)  2265--2272

\bibitem{kae2013augmenting}
Kae, A., Sohn, K., Lee, H., Learned-Miller, E.:
\newblock Augmenting crfs with boltzmann machine shape priors for image
  labeling.
\newblock In: Proceedings of the IEEE Conference on Computer Vision and Pattern
  Recognition. (2013)  2019--2026

\bibitem{chen2014detect}
Chen, X., Mottaghi, R., Liu, X., Fidler, S., Urtasun, R., Yuille, A.:
\newblock Detect what you can: Detecting and representing objects using
  holistic models and body parts.
\newblock In: CVPR. (2014)

\bibitem{sagonas2013300}
Sagonas, C., Tzimiropoulos, G., Zafeiriou, S., Pantic, M.:
\newblock 300 faces in-the-wild challenge: The first facial landmark
  localization challenge.
\newblock In: International Conference on Computer Vision, (ICCV-W), 300 Faces
  in-the-Wild Challenge (300-W), Sydney, Australia, 2013, IEEE (2013)

\bibitem{belhumeur2011localizing}
Belhumeur, P., Jacobs, D., Kriegman, D., Kumar, N.:
\newblock Localizing parts of faces using a consensus of exemplars.
\newblock In: CVPR. (2011)

\bibitem{le2012interactive}
Le, V., Brandt, J., Lin, Z., Bourdev, L., Huang, T.S.:
\newblock Interactive facial feature localization.
\newblock In: ECCV. (2012)

\bibitem{ramanan2011}
Zhu, X., Ramanan, D.:
\newblock Face detection, pose estimation, and landmark estimation in the wild.
\newblock In: CVPR. (2012)

\bibitem{jia2014caffe}
Jia, Y., Shelhamer, E., Donahue, J., Karayev, S., Long, J., Girshick, R.,
  Guadarrama, S., Darrell, T.:
\newblock Caffe: Convolutional architecture for fast feature embedding.
\newblock arXiv preprint arXiv:1408.5093 (2014)

\end{thebibliography}
\end{document}